\documentclass[pdflatex,sn-mathphys-num]{sn-jnl}
\usepackage{xcolor}
\usepackage{graphicx}
\usepackage{multirow}
\usepackage{amsmath,amssymb,amsfonts}
\usepackage{amsthm}
\usepackage{mathrsfs}
\usepackage[title]{appendix}
\usepackage{xcolor}
\usepackage{textcomp}
\usepackage{manyfoot}
\usepackage{booktabs}
\usepackage{algorithm}
\usepackage{algorithmicx}
\usepackage{algpseudocode}
\usepackage{listings}
\usepackage{subcaption}
\usepackage{bm}
\usepackage[capitalise]{cleveref}
\usepackage{lineno}

\newcommand{\rev}[1]{\textcolor{black}{#1}}

\newcommand{\Tws}{\mathsf{T}^{W}_{S}}

\newcommand{\myQ}{\mathsf{Q}}

\newcommand{\conic}{\mathsf{E}_{ij}}

\newcommand{\myP}{\mathsf{P}}

\DeclareMathOperator*{\argmin}{arg\,min}

\theoremstyle{thmstyleone}

\theoremstyle{thmstyletwo}

\theoremstyle{thmstylethree}

\begin{document}
\title[Article Title]{Acquiring Submillimeter-Accurate Multi-Task Vision Datasets for Computer-Assisted Orthopedic Surgery}



\author*[1,2]{Emma Most} \email{emmost@ethz.ch}
\author[1,2]{Jonas Hein}
\author[1]{Frédéric Giraud}
\author[1]{Nicola A. Cavalcanti}
\author[1]{Lukas Zingg}
\author[3]{Baptiste Brument}
\author[1]{Nino Louman}
\author[1]{Fabio Carrillo}
\author[1]{Philipp Fürnstahl}
\author[1]{Lilian Calvet}

\affil[1]{Research in Orthopedic Computer Science, University Hospital Balgrist, University of Zurich, Switzerland}
\affil[2]{Computer Vision and Geometry, ETH Zurich, Switzerland}
\affil[3]{Institut de Recherche en Informatique de Toulouse, France}

\abstract{\textbf{Purpose:} Advances in computer vision, particularly in \rev{optical image-based} 3D reconstruction and feature matching, enable applications like marker-less surgical navigation and digitization of surgery. However, their development is hindered by a lack of suitable datasets \rev{with 3D ground truth}. This work explores an approach to generating realistic and accurate \rev{\textit{ex vivo}} datasets tailored for 3D reconstruction and feature matching in open orthopedic surgery. \\
\textbf{Methods:} A set of posed images and an accurately registered ground truth \rev{surface mesh} of the scene are required to develop vision-based 3D reconstruction and matching methods suitable for surgery. We propose a framework consisting of three core steps and compare different methods for each step: 3D scanning, calibration of viewpoints for a set of high-resolution RGB images, and an optical-based method for scene registration.\\
\textbf{Results:} We evaluate each step of this framework on an \textit{ex vivo} scoliosis surgery using a pig spine, conducted under real operating room conditions.  \rev{A mean 3D Euclidean error of 0.35 mm is achieved with respect to the 3D ground truth.} \\
\textbf{Conclusion:} The proposed method results in submillimeter accurate 3D ground truths and surgical images with a spatial resolution of 0.1 mm. This opens the door to acquiring future surgical datasets for high-precision applications.}

\maketitle

\section{Introduction}
Computer vision tasks are widely used in orthopedic surgery for various applications, including surgical navigation \cite{7DSurgical2021}, robotic-assisted surgery \cite{Kiyasseh2022}, and the creation of surgical digital twins \cite{Hein_2024}.
Computer vision enables real-time alignment of intraoperative optical images with preoperative 3D models of the anatomy \cite{florentin}, facilitating precise navigation of anatomical structures, including hidden substructures, for both surgeons and robotic systems.
3D reconstruction of the anatomy and feature matching are examples of typical tasks required by computer-assisted orthopedic surgery (CAOS) systems.
Accurate solutions to these tasks have the potential to eliminate the need for markers, which are associated with a complex workflow \cite{haertl}.
Surgical digital twins also benefit from advances in 3D reconstruction, allowing for high fidelity replica of real-world surgery. These 3D reconstructions can for example be used for education, where they can provide medical students and surgeons with interactive and virtual environments, and to train surgical robots in highly realistic simulations \cite{RoboticSurgeryRL2021}. 

The development of these computer vision methods requires large, realistic and surgical datasets \rev{with accurate 3D ground truths}. While extensive datasets exist for man-made environments \cite{Lee2024}, the medical field lags behind due to ethical and logistical challenges. Existing surgical datasets focus on minimally invasive surgery (MIS), and available open surgery datasets lack the realism and accuracy needed for precision applications \cite{tracking_review}. This work addresses these gaps by working towards a method to acquire realistic \rev{\textit{ex vivo}} datasets with highly accurate 3D ground truth \rev{of the anatomy, represented as a surface mesh, and optical images with precise corresponding camera poses.}

The contributions of this work are a comparative analysis of methods for acquiring an accurate surface mesh of the \rev{visible} anatomy, a comparison between different calibration techniques to obtain camera poses, and a marker-based method for registering the surface mesh with the posed images, together with a method to assess the accuracy of each of these steps. Each proposed step yields a very high accuracy and therefore, our work promises significant potential for \rev{capturing realistic \textit{ex vivo} surgical datasets}. We also provide a pilot dataset, validated using a pig torso to simulate scoliosis surgery and use our dataset to evaluate state-of-the-art (SOTA) \rev{surface} reconstruction methods in sparse or dense viewpoint scenarios. \rev{The code and dataset can be found under \href{https://github.com/emmamost26/CamSceneRegistration}{https://github.com/emmamost26/CamSceneRegistration}.}

\section{Related Work}
\noindent \textbf{Anatomy \rev{surface} reconstruction}
\rev{CT and MRI, while excellent for preoperative imaging, are challenging and impractical for intraoperative 3D reconstruction. CT exposes patients to ionizing radiation, making repeated use undesirable, and MRI requires a magnetic field-free environment, limiting compatibility with standard surgical tools. Moreover, both modalities lack real-time imaging capabilities and are not typically available in operating rooms, adding logistical and cost challenges.
Ultrasound (US), though real-time, is operator-dependent, and requires direct contact with the anatomy.
In contrast, optical cameras are the preferred solution for 3D reconstruction of the visible anatomy. They provide real-time, radiation-free imaging and capture anatomical details without requiring physical contact.}

\rev{Optical image-based} methods like Structure from Motion (SfM) and Simultaneous Localization and Mapping (SLAM) have been adapted for surgical applications, with some focusing on endoscopic images to map and track anatomy in real time \cite{mahmoud2017orbslam}. Deep learning-based approaches, such as neural radiance fields (NeRF) and transformer-based stereoscopic depth perception, have shown improved results in surgical scene reconstruction \cite{dl_3dreconstruction_survey}. Structured light techniques have also been explored but are less suitable for real-time applications due to narrow depth of field and acquisition time \cite{KellerA00}. Despite progress, most 3D reconstruction methods focus on MIS, highlighting a gap in open surgery methods that this work aims to address. 

\noindent \textbf{Anatomy tracking}
Marker-less tissue tracking, primarily explored in endoscopic surgery, often registers a preoperative \rev{surface mesh} with early intraoperative data. Notable works include \cite{Richa2008}, which tracked heart motion in MIS, and \cite{Yip2012}, which used stereo-cameras for real-time tissue tracking during partial nephrectomy. 
 
In open surgery, \cite{florentin} developed a method for marker-less registration of preoperative lumbar spine models using RGB-D data from an overhead stereo camera. Despite its promise, the method’s accuracy is limited by the unrealistic cadaveric dataset used, which differs from actual surgical conditions. Improving precision necessitates the collection of more realistic datasets as concluded in \cite{tracking_review}.

\begin{figure}[t]
    \centering
    \includegraphics[width=0.8\textwidth]{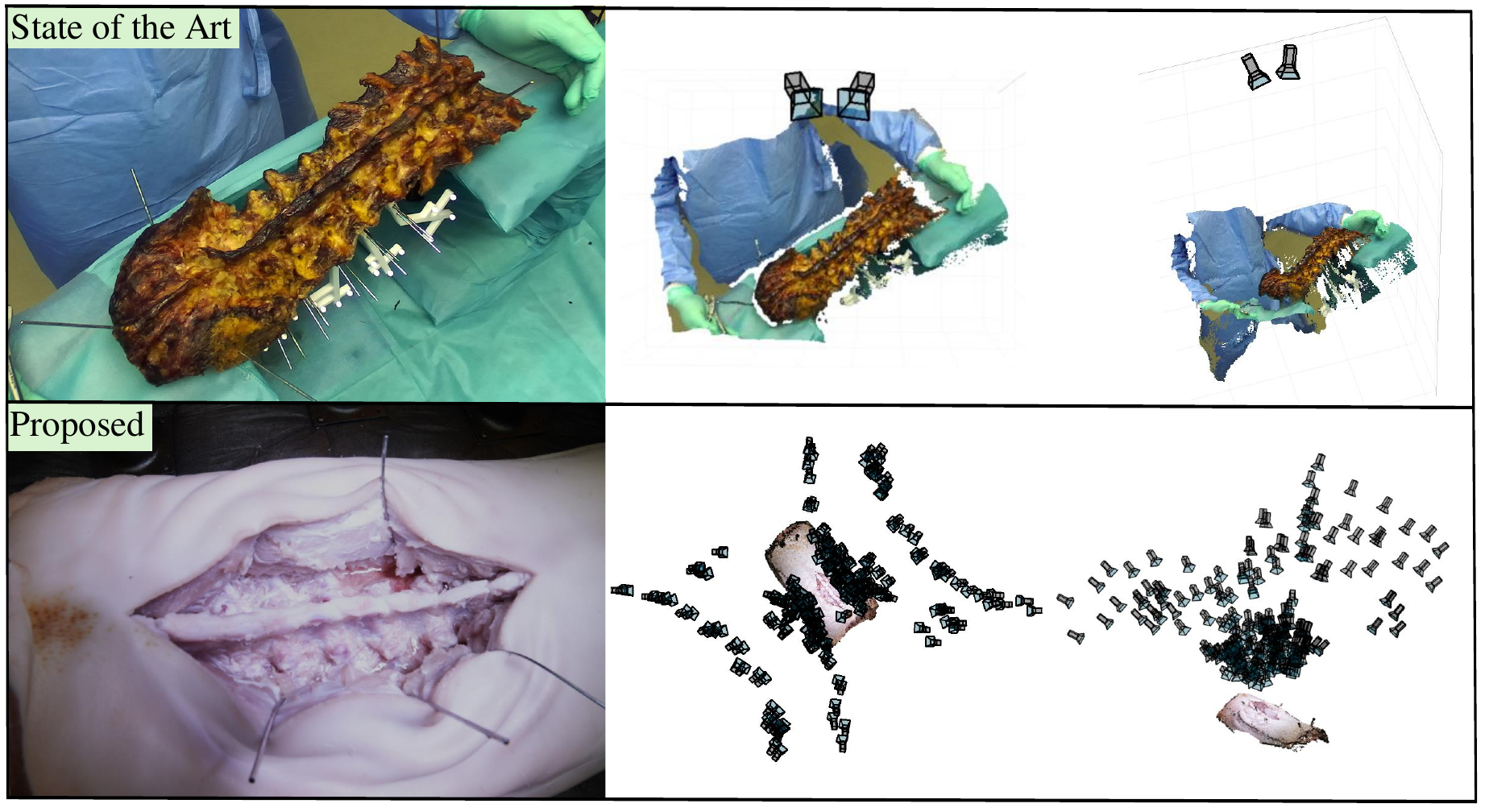}
    \caption{Comparison of our dataset acquisition method compared to SpineDepth \cite{spinedepth}. SpineDepth offers limited viewpoint diversity (2 camera poses), cadaver images that are unrealistic for surgery, a mean target registration error of 1.5 mm, and a median deviation between ground truth and measured anatomy of 2.4 mm. Our method allows for unlimited viewpoints (216 were taken for experiments), realistic images, with a mean radial registration error of 0.35 mm.
}
    \label{fig:related_work} 
\end{figure}

\noindent \textbf{Datasets with 3D ground truth} 
\rev{Large datasets containing posed images and 3D ground truths of indoor and outdoor man-made environments \cite{omnidata, megadepth} are a crucial prerequisite for data-driven 3D reconstruction \cite{yu2022monosdf, dust3r} and feature matching \cite{lindenberger2023lightglue} methods, which have demonstrated clear superiority compared to traditional approaches \cite{dust3r, Sarlin_2020_CVPR}.} Recent trends in MIS have also pushed the publication of endoscopic datasets amongst which some provide 3D ground truth and annotated poses \cite{endomapper} and are thus also suitable for surface reconstruction. However, datasets featuring only man-made scenes do not address the complexities of surgical data and MIS datasets are unsuitable for open surgery due to anatomical differences and lower image quality. 

To the best of our knowledge, the only existing open surgery dataset is SpineDepth \cite{spinedepth}, which provides posed RGB-D images and the 3D scene geometry of dissected lumbar spines. 
However, the reported ground-truth accuracy of 1.5 mm is insufficient for the training of pixel-accurate feature matching methods or high-quality surface reconstructions. 
Further limitations include the unrealistic exposure of the anatomy and a very limited number of viewpoints, as shown in \cref{fig:related_work}. 
In contrast, our work proposes a methodology for the automated capture of high-quality images with submillimeter accuracy of both camera poses and the surface mesh of the anatomy. 
The method is designed specifically for surgical applications, supporting the development and evaluation of marker-less 3D reconstruction and tracking techniques.

\section{Methodology}
In this section, we describe our proposed acquisition method to collect \rev{an accurate surface mesh of the scene registered to posed images with sub-millimeter accuracy. By \textit{scene}, we refer to the specimen placed on an operating table, along with a set of 3D markers, consisting of spheres with known radii, fixated around it. }

\rev{Our method comprises three steps, namely the scene surface reconstruction (Section \ref{ssec:methods_scanning}), the capture of posed images (Section \ref{ssec:methods_pose_estimates}), and the registration of the posed images with the surface of the scene (Section \ref{ssec:methods_scene_registration}). 
Separating the acquisition process into these three steps provides modularity and enables the comparison of state-of-the-art solutions for each step.}

\subsection{Scene Surface Reconstruction}
\label{ssec:methods_scanning}
CT scanning is a well-established gold standard for acquiring 3D ground truth models of anatomical structures. 
Modern CT scanners achieve high spatial resolutions, making them ideal for capturing intricate anatomical details. 
For our CT baseline, we perform a CT scan on the animal specimen with a spatial resolution of 0.4mm$^3$ (NAEOTOM Alpha, Siemens, Germany). 
The anatomy is segmented using Mimics (Materialise, Leuven, Belgium), followed by the extraction of a surface mesh.

However, CT scanning comes with several limitations: it is costly, not always easily accessible, and presents logistical challenges. 
Additionally, for the capture of an annotated dataset, transporting the anatomy between a wet lab or an operating room to an imaging center can introduce deformation, compromising the accuracy of the dataset.
These limitations motivated us to compare CT to optical scanning, which eliminates the need to move the anatomy during the data capture. 
In this study, we utilize the \textit{Space Spider} handheld 3D scanner (Artec 3D, Luxembourg), which offers a high point accuracy of up to 0.05 mm, and a spatial resolution of 0.1 mm, making it a promising alternative to CT scanning. \rev{Note that the scene is scanned such that the positions and geometries of the markers are captured in the mesh. These are then used for scene registration, as described in \cref{ssec:methods_scene_registration}.}

\subsection{Capturing Posed Images}
\label{ssec:methods_pose_estimates} 
\rev{Data capture can be performed manually or using a robotic arm. Manual capture requires minimal hardware and is most versatile, while mounting the camera on a robotic arm allows for automation. This second option is chosen for its scalability in surgical \textit{ex vivo} data captures.}

\rev{Camera poses can be obtained either using SfM \cite{colmap, glomap} or, if a robotic arm is used, using the robot's forward kinematics. We evaluate these two approaches for camera pose estimation.}
\rev{The camera pose estimation based on the robot's forward kinematics involves determining} the transformation between the camera $C$ and the robot's end-effector $EE$, referred to as $T_{C}^{EE} \in \mathbb{R}^{4 \times 4}$ in the sequel. The camera pose can be expressed in the fixed coordinate frame of the robot's base $B$ as
\begin{equation}
    T_{C}^{B} = T_{EE}^{B} \cdot T_{C}^{EE},
\end{equation}
where $T_{EE}^{B}$ is the Euclidean transformation from the robot's end-effector to base coordinate frame.
The calibration of $T_{C}^{EE}$ is detailed in the supplementary material.

To enable a fair comparison between the SfM and robot-based camera pose estimation approaches, we evaluate both approaches on the same set of images captured with the camera attached to the robot arm.

\subsection{Scene Registration}
\label{ssec:methods_scene_registration}
\begin{figure}[t]
    \centering
    \includegraphics[width=\linewidth]{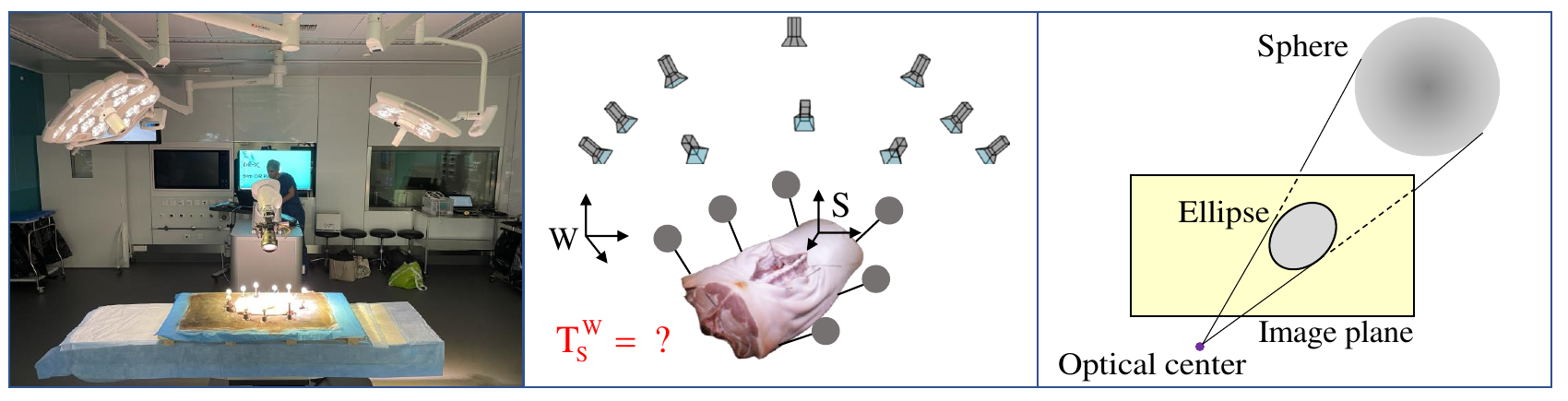}
    \caption{\textbf{Left:} Scene registration on a pig spine surgery in real operating conditions. \textbf{Middle:} Given a set of $N$ posed images of the scene expressed in the world reference frame $W$ and a surface mesh of the scene expressed in a local reference frame $S$, the scene registration consists in recovering the relative pose $\mathsf{T}_S^W$. We rigidly attach $M$ spherical markers to the scene and fit spheres to \rev{the corresponding regions in the surface mesh to estimate their positions in $S$.} \textbf{Right:} The spherical markers project into the image as ellipses, which are automatically detected and used to recover $\mathsf{T}_S^W$.}
    \label{fig:scene_registration}
\end{figure}

The final step involves registering the posed images with the \rev{surface mesh} using 3D printed spherical optical markers rigidly affixed around the anatomy (see Figure \ref{fig:scene_registration}). \rev{These spheres are precisely localized in the surface mesh by fitting virtual spheres of the same radius to the mesh vertices using the Iterative Closest Point (ICP) algorithm, with manual initialization.} 
The precisely known positions and geometry of the markers make them reliable for accurate registration, compensating for the inherent limitations of SfM, which typically produces poses on a non-metric scale. 
Note that these markers are used solely for scene registration, not for camera pose estimation. 

Inspired by \cite{zhang2007Ellipse}, we perform a target-based registration from images of spheres to have the surface mesh of the scanned scene and the posed images expressed in a common reference frame. Suppose a set of posed images of a scene expressed in world reference frame $W$ and a surface mesh of the scene expressed in the local reference frame attached to the scene $S$. The goal is to determine the relative pose $\Tws \rev{\in \mathbb{R}^{4 \times 4}}$.
\\\\
\noindent\textbf{Optimization Objective} 
\rev{
A sphere can be represented as a quadric matrix $\mathsf{Q} \in \mathbb{R}^{4 \times 4}$, which is symmetric and defined as a function of its radius and center $\mathbf{c}_\mathsf{W} = \Tws \mathbf{c}_\mathsf{S}$.
It projects into the image as an ellipse $\mathsf{E}$ of equation:}
\begin{equation}
\mathsf{E}^{-1} = \myP \myQ^{-1} \myP^T,
\end{equation}
\rev{where $\myP \in \mathbb{R}^{3 \times 4}$ is the camera projection matrix. Let $\mathbf{x}$ denote the augmented Cartesian coordinates of a point on the ellipse in the 2D image plane. Any such point satisfies:}
\begin{equation}
\mathbf{x}^T \mathsf{E} \mathbf{x} = 0 .
\end{equation} We take this bilinear product as the cost to our minimization problem and solve it using the Levenberg–Marquardt optimization method. Summing this cost over all $N$ images, $M$ markers and a chosen number of $\rev{L}$ points on the ellipse, the minimization can be written as follows: 
\begin{equation}
    \argmin_{\rev{\Tws}} \sum_{i=1}^{N}\sum_{j=1}^{M}\sum_{l=1}^{L}\lVert \mathbf{x}_{ijl}^T \conic \mathbf{x}_{ijl} \rVert_2^2
\label{eq:registration_objective}
\end{equation}
\rev{where $\Tws$ represents the rigid transformation from the local coordinate frame $S$ of the surface mesh to the coordinate frame $W$, in which the posed images are expressed. When the input poses are computed using SfM, a scale factor is jointly estimated for the camera poses.}
The extraction of $\rev{L}$ points on the ellipse outline is detailed in the supplementary material.
\\\\
\noindent\textbf{Initial Estimates}
\rev{The initial estimate for $\Tws$ is computed with a Perspective-n-Point (PnP) solver,} using the centers of the ellipses and corresponding centers of the 3D spheres as 2D-3D correspondences in one image of the image collection for which all the markers are well visible.

\rev{Matching each imaged sphere to its corresponding 3D sphere is performed by exhaustively solving the PnP problem for all combinations of four selected ellipse centers paired with the $M$ 3D sphere centers.}

\section{Experiments and Results}
\label{sec:results}

\begin{figure}[t]
    \centering
    \begin{minipage}{0.28\textwidth}
        \centering
        \includegraphics[width=\textwidth]{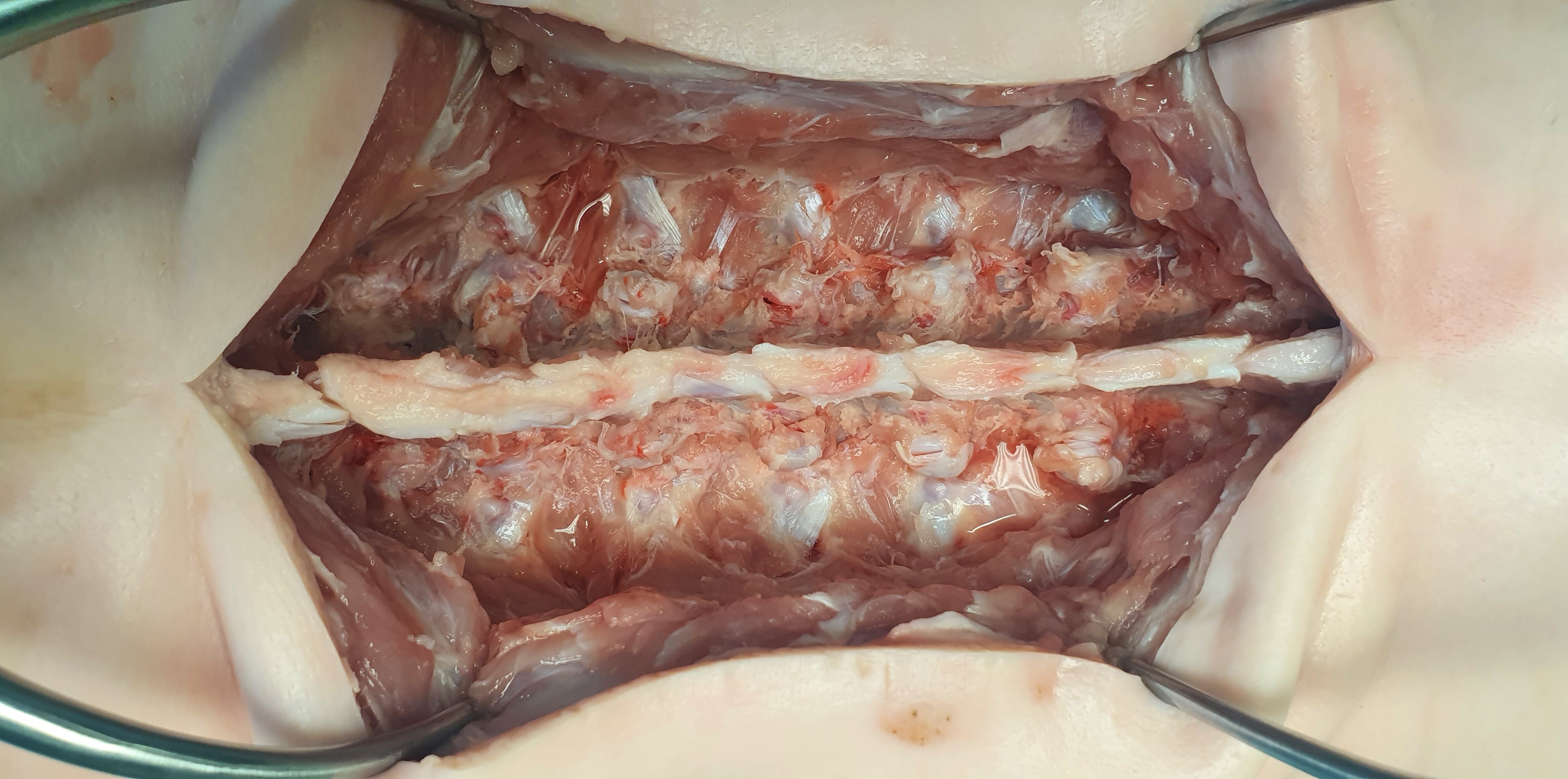}
    \end{minipage}
    \begin{minipage}{0.28\textwidth}
        \centering
        \includegraphics[width=\textwidth]
        {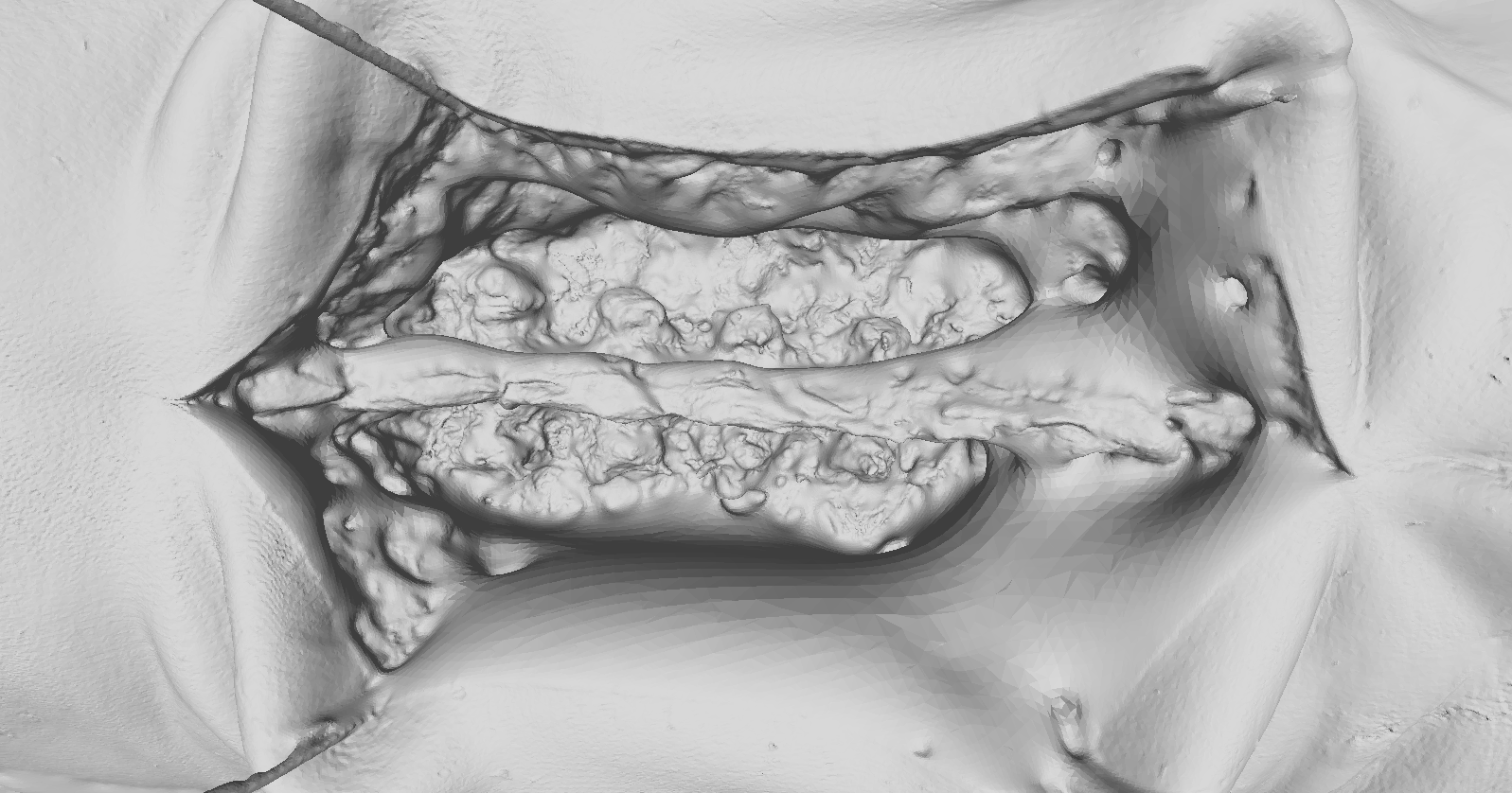}
    \end{minipage}
    \begin{minipage}{0.28\textwidth}
        \centering
        \includegraphics[width=\textwidth]{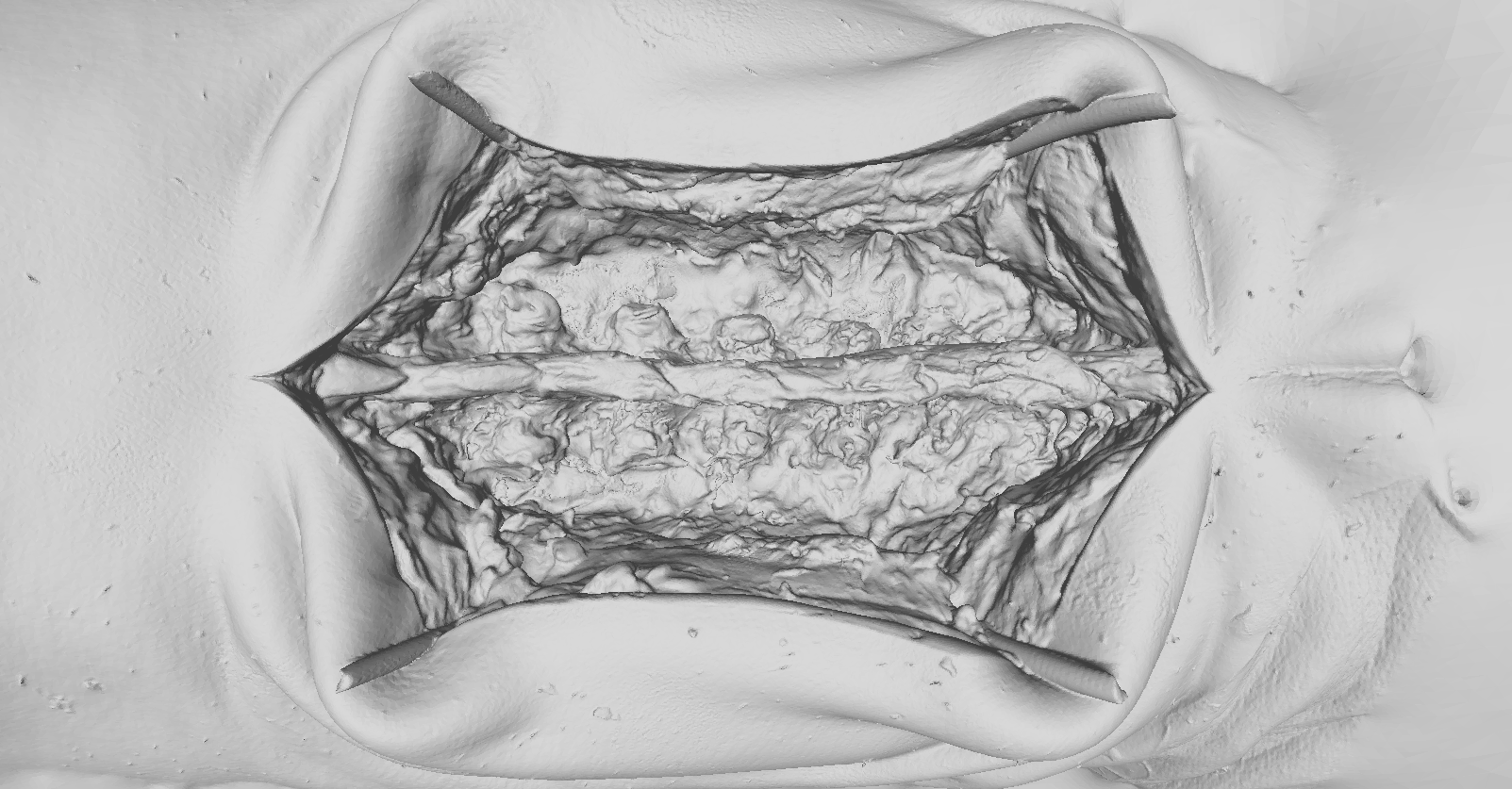}
    \end{minipage}
    \caption{\textbf{Left:} RGB image of the open spine of a pig. \textbf{Middle:} Reconstruction without coating. \textbf{Right:} Reconstruction with coating. The reconstruction with the spray clearly captures more detail, demonstrating the benefit of using the spray for improved surface reconstruction.}
    \label{fig:scanning_spray}
\end{figure}

\subsection{Acquisition Protocol}
We evaluated our proposed methodology in a simulated \textit{ex vivo} scoliosis surgery using a pig spine. \rev{The specimen was rigidly fixed onto a wooden board with K-wires. An incision mimicking a scoliosis surgery} was made by a clinician and held open using additional K-wires. \rev{3D-printed spherical markers (30 mm diameter) were affixed to a wooden board. Before collecting optical image data, the setup was transported to the imaging center for CT scanning. Afterwards, it was returned to the operating room, placed on the operating table, and positioned alongside a robotic arm (LBR Med 14 surgical robot arm, KUKA AG, Germany) with a high-resolution camera (Alpha 7R V with a FE 24-70 mm F2.8 GM lens, Sony Group Corporation, Tokio, Japan) mounted on its end-effector. The camera was focused on the scene's center and its internal calibration was performed.
Subsequently, $N=108$ images were captured from two robot positions on opposite sides of the operating table, including 30 viewpoints specifically selected to ensure good marker visibility for the scene registration. Note that the dataset images do not necessarily contain the markers, ensuring that they accurately represent a realistic human surgery scene. Images with visible markers can be cropped to remove them. Due to the high initial resolution (9504x6336 px), cropped images retain a sufficiently high resolution and optimal realism, as highlighted in the supplementary material.}

\rev{Following image acquisition, Aesub Blue scanning spray (AESUB GmbH, Germany) was applied to the anatomical surface, similar to the approach described in \cite{skoltech3d}, to reduce reflectivity and enhance scanning quality (see \cref{fig:scanning_spray}). Finally, the entire scene, including the markers, was scanned with the optical scanner.}

\subsection{Scene \rev{Surface} Reconstruction}

\begin{figure}[t]
    \centering
    \includegraphics[width=0.99\linewidth]{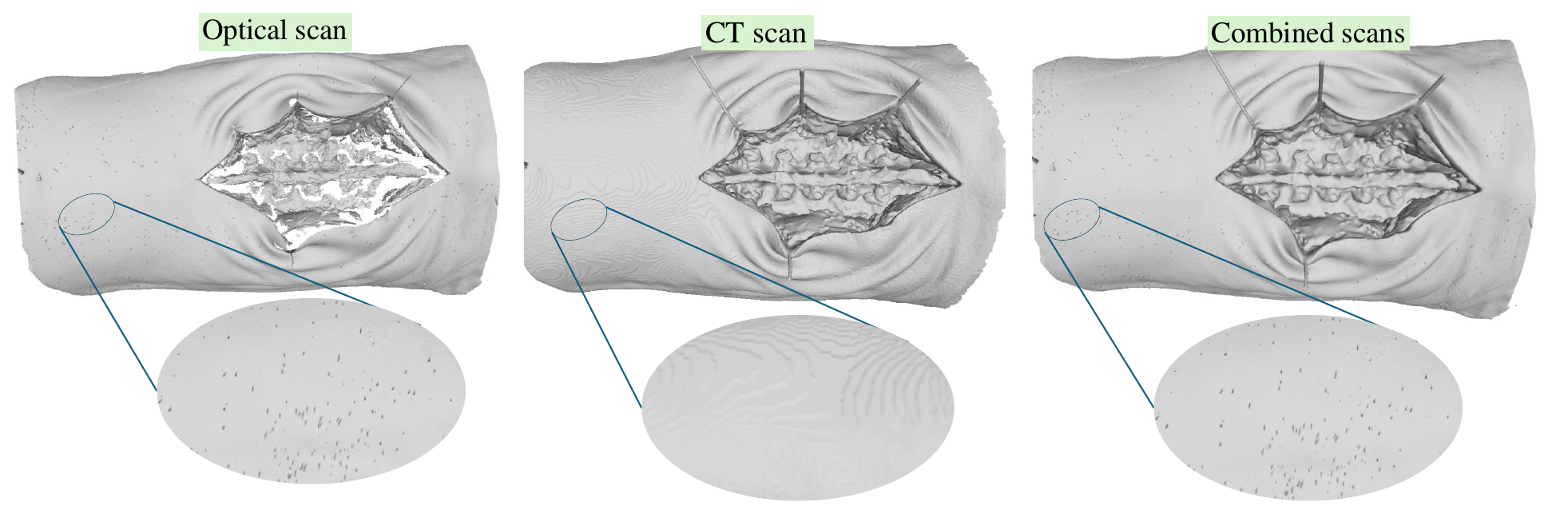}
    \caption{\textbf{Left:} The optical scan obtained from the Artec3D Space Spider is able to reconstruct very small details such as the specimen's hair. \textbf{Middle:} However, the CT scan with resolution 0.4x0.4x0.4 mm performs better on concave parts such as the interior of the wound. \textbf{Right:} Combining the two methods—using the CT scan for concave areas and the optical scanner for the rest—yields the best results.}
    \label{fig:scan_comparisons}
\end{figure}
\begin{figure}[t]
    \centering
    \includegraphics[width=\linewidth]{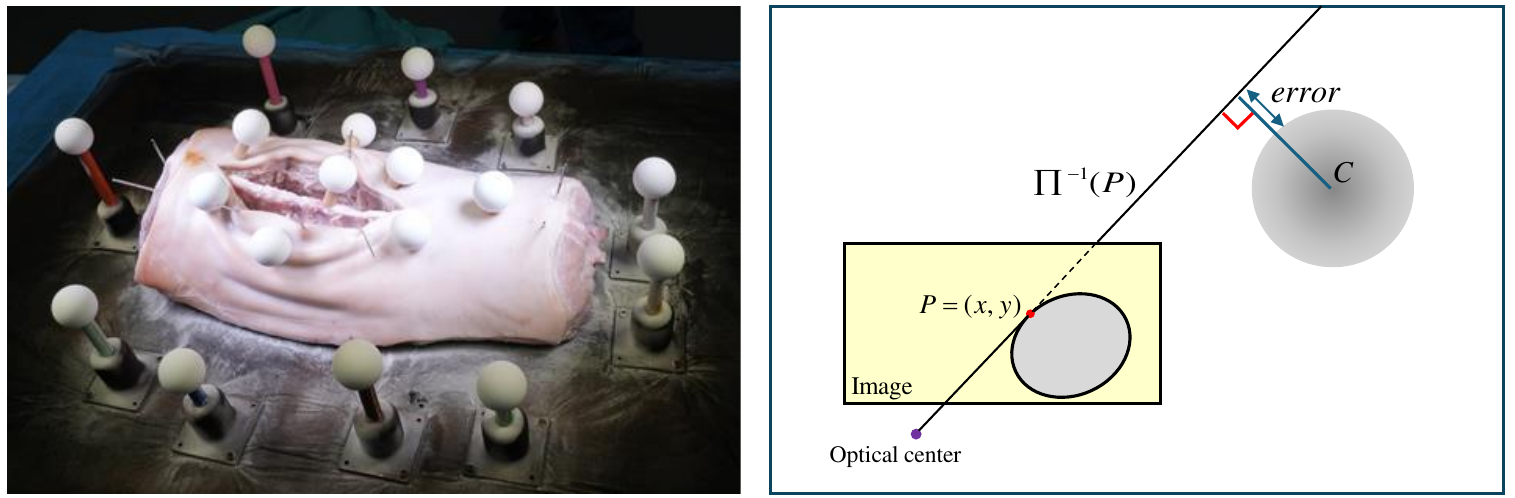}
    \caption{\textbf{Left:} Scene registration evaluation. \rev{Control markers} were placed at the center of the scene. \textbf{Right:} We define the radial error to be the Euclidean distance between a ray back-projected from a point on the outline of the ellipse corresponding to an evaluation marker and the hull of the marker in 3D space.}
    \label{fig:evaluation_setup}
\end{figure}

We provide a visual comparison of the reconstructions from the CT scan and the optical scan in Figure \cref{fig:scan_comparisons}.
Low-frequency geometric features were accurately reconstructed with an average Chamfer distance of 0.7 mm between the optical scan and the CT-based model.
High-frequency details were also well captured \rev{ by the optical scan}, as shown qualitatively in Figure \ref{fig:scan_comparisons}. 
While a CT scan of even higher spatial resolution might recover these details, the handheld optical scanner proved to be an excellent alternative. 
One drawback of the handheld scanner, however, is its partial reconstruction of concave regions, such as inside deep incisions.

We draw the following conclusions: (i) applying a coating is highly beneficial for improving scan quality; (ii) the optical scanner is a suitable and efficient alternative to CT for scanning anatomical geometries, except for deep concavities like wounds. For such regions, CT scanning remains the preferred option.

Finally, the surface meshes from CT scan and optical scanner can optionally be fused to obtain the optimal result in both concave and convex areas. \rev{To achieve this, one can either perform a 3D-to-3D registration of the spheres, as they are present in both scans and exhibit similar scanning quality, or run the proposed image-based scene registrations twice: one for the optical scan mesh and another for the CT mesh. In our experiment, we opted for the latter approach, aligning both meshes to a common coordinate frame, specifically that of the cameras.
We then manually segment the CT scan mesh to retain only the wound region and the optical scan mesh to retain only its outer region, while ensuring a few millimeters of overlap to avoid gaps in the final mesh. The two registered regions are then merged into a single mesh, resulting in the final ground truth mesh expressed in the cameras' frame. This approach was used to create the pilot dataset, producing the anatomical mesh shown in the right image of \cref{fig:scan_comparisons}}.

\subsection{\rev{Camera Poses and Scene Registration}}
\label{ssec:results_scene_registration}

\begin{figure}[t]
    \centering
    \includegraphics[width=\linewidth]{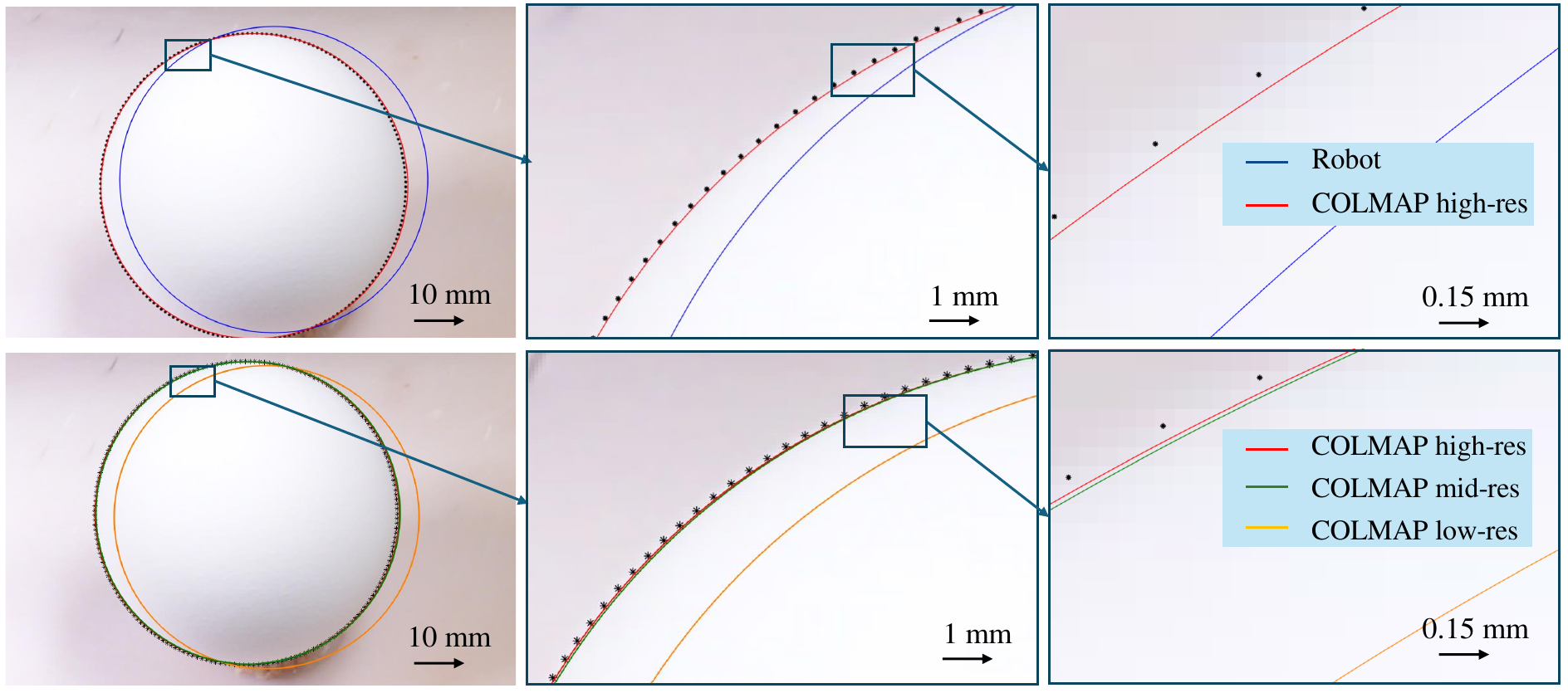}
    \caption{Qualitative comparison of the estimated camera poses from different modalities. \textbf{Top:} Using COLMAP with high-res images vs using robot end-effector poses. \textbf{Bottom:} Using COLMAP with different image resolutions (resp. 9504x6336 px, 4752x3168 px and 1920x1080 px). We obtained a mean reprojection error of 3.71 px using high-res camera poses from COLMAP against 8.99 px using robot poses. COLMAP with mid-res and low-res poses reported reprojection errors of resp. 3.91 px and 11.53 px.}
    \label{fig:registration_results}
\end{figure}

To evaluate the scene registration, we positioned spherical \textit{control markers} on top of the specimen and inside the incision, distinct from the M = 10 \textit{registration markers} which were placed around the specimen (see \cref{fig:evaluation_setup}). 
We captured N = 16 images from different viewpoints, scanned the scene with the optical scanner, and extracted the 3D marker locations, similarly to what was done in \cref{ssec:methods_scene_registration}). 
For each evaluation marker, we sampled points on the corresponding ellipse outline \rev{ observed on each image} and defined the radial error as the distance between a back-projected ray from a point on the ellipse and the corresponding 3D sphere, as depicted in \cref{fig:evaluation_setup}. 
Reprojection error (in pixels) was calculated between the ellipse obtained by projecting the sphere into the image with the estimated scene registration solution and detected ellipse. 
We compare the results across robot, COLMAP \cite{colmap}, and GLOMAP \cite{glomap} poses in \cref{table:registation_errors} and \cref{fig:registration_results}. \rev{Our analysis shows that using COLMAP poses with either high resolution (9504x6636 px) or medium resolution (4752x3168 px) images yields the most accurate results. While the accuracy with high-resolution images is slightly superior, the difference is minimal compared to the results obtained with medium-resolution images.}

While using robot poses has the advantage of being independent of the scene appearance, it highly relies on the accuracy of the poses delivered by the robot and implies that the robot base must stay fixed during the entire data acquisition, which is inconvenient to capture data from all angles.

\begin{table}[t]
\centering
\setlength{\tabcolsep}{8pt} 
\renewcommand{\arraystretch}{1}
\begin{tabular}{lccccc}
\hline
                             & Robot   & \multicolumn{3}{c}{COLMAP}            & GLOMAP  \\
Image Resolution             & -       & Low      & Medium  & High             & High    \\ \hline
Mean radial error (mm)       & 0.89    & 1.10     & 0.37    & \textbf{0.35}    & 0.44    \\
Mean reprojection error (px) & 8.99    & 11.53    & 3.91    & \textbf{3.71}    & 4.67    \\ \hline
\end{tabular}
\caption{The scene registration method is evaluated in terms of radial errors and mean reprojection error (detailed in \cref{ssec:results_scene_registration}) for camera poses obtained from the robot, COLMAP, or GLOMAP. 
For the SfM approaches we compare the registration errors when using low (1920x1080 px), medium (4752x3168 px) and high-resolution (9504x6336 px) images.}
\label{table:registation_errors}
\end{table}

\begin{figure}[t]
    \centering
    \begin{subfigure}[b]{0.2\textwidth}
        \centering
        \footnotesize
        \raisebox{1.25cm}{
        \begin{tabular}{lcccc}
            & \multicolumn{4}{c}{Chamfer distance (mm) $\downarrow$} \\
            \midrule
             & COLMAP & Instant-NGP & NeuS2 & SuGaR \\ 
            \midrule
            Dense & \textbf{0.68} & 2.31 & 1.23 & 2.99\\ 
            Sparse & - & \textbf{3.89} & 4.27 & 5.35\\
        \end{tabular}}
        \label{fig:table}
    \end{subfigure}
    \hfill
    \begin{subfigure}[b]{0.4\textwidth}
        \centering
        \includegraphics[width=\textwidth]{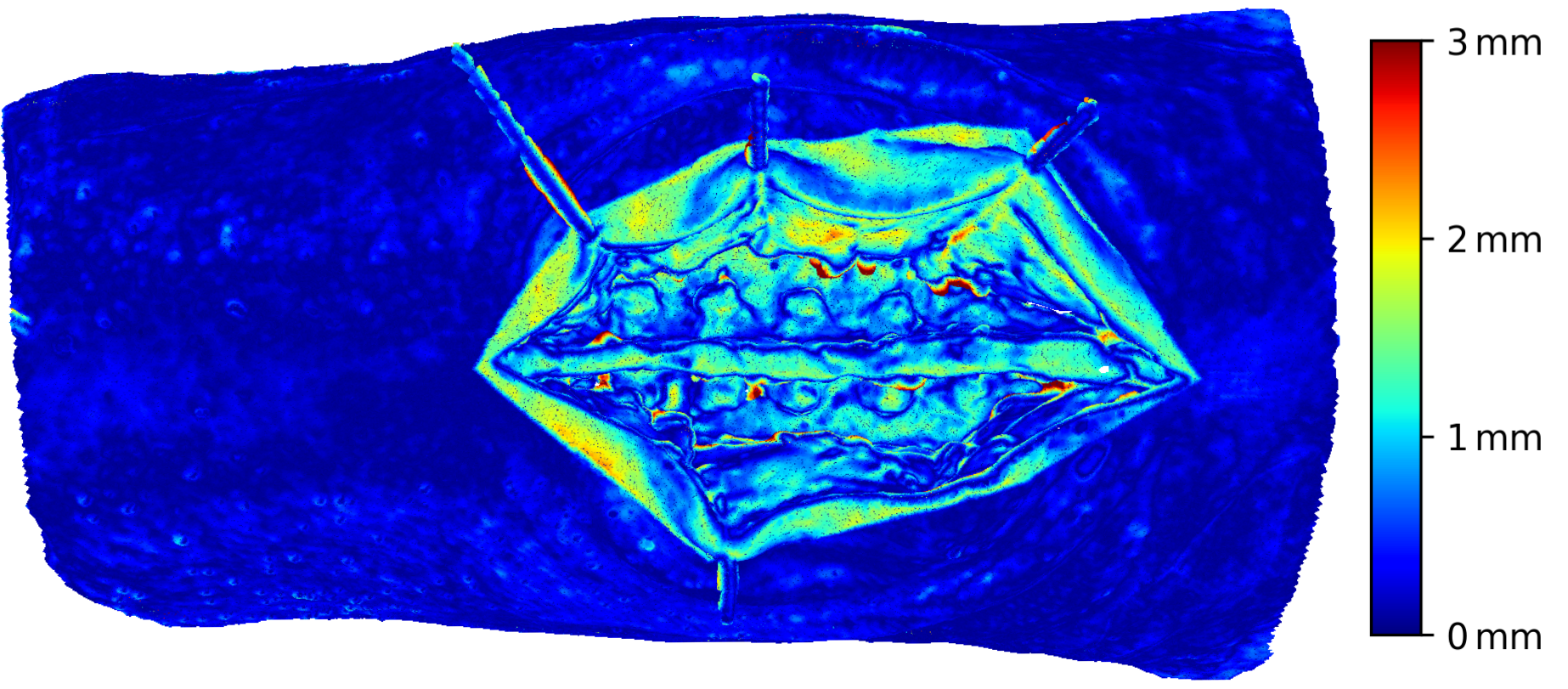}
        \label{fig:plot}
    \end{subfigure}
    
    \caption{\textbf{Left}: surface reconstruction errors (mm), measured by Chamfer distance (lower is better), are compared across two acquisition scenarios: (i) dense viewpoints (N=216) at a resolution of 3840x2160 pixels, and (ii) sparse viewpoints (N=8) at 1920x1080 pixels. COLMAP performs best in the dense viewpoint scenario but fails to reconstruct the scene with only N=8 viewpoints, where Instant-NGP shows the best performance. \textbf{Right}: Heatmap showing the Chamfer distance between the model reconstructed with COLMAP and the proposed ground truth.}
    \label{fig:main}
\end{figure}

\subsection{\rev{Application to} 3D reconstruction}

We demonstrate one of the uses of our pilot dataset as a benchmark to compare different methods of 3D reconstruction \rev{from posed images}. 
For this, we tested four methods which reconstruct the surface of a scene from RGB data, namely \rev{the traditional multi-view stereo method} COLMAP \cite{colmap}, \rev{the NeRF-based methods} Neus2 \cite{neus2} and Instant-NGP \cite{mueller2022instant}, and \rev{the Gaussian splatting-based method} SuGaR \cite{guedon2023sugar}.
We evaluate them against our ground truth surface mesh. \rev{Note that the posed images used for the 3D reconstructions, as well as the 3D model serving as ground truth (\cref{fig:scan_comparisons}, right), are all expressed in the same coordinate frame. Consequently, the resulting 3D reconstruction can be directly compared to the proposed ground truth mesh.} We evaluated the methods in a sparse scenario on mid-resolution images, using a subset (N=8) of our captured images, simulating the use case of surgical navigation, where only few cameras are typically placed around the anatomy. We also evaluated them in a dense scenario (N=216) using high-res images, which would correspond to the use-case of digitization \rev{ using \textit{ex vivo} specimens}. Results are presented in Figure \ref{fig:main}. \rev{COLMAP achieves the best performance in the dense scenario}, with a Chamfer distance to our ground truth of 0.68 mm, but fails in the sparse scenario. \rev{Instant-NGP, however, holds in the sparse scenario, with a Chamfer distance of 3.89 mm}, outperforming the other evaluated methods. \rev{Details on the computation of the Chamfer distance between the predicted meshes and the ground truth mesh are provided in the supplementary material.}

\section{Conclusion}
We proposed a framework to acquire surgical datasets comprising an accurate surface mesh of the scene and posed images intended for the development and benchmarking of 3D reconstruction and feature matching methods. 
We evaluated various approaches for 3D scanning, recovering camera poses, and registering the scene along with the camera poses in an \textit{ex vivo} scoliosis surgery experiment using a pig spine, conducted under real operating conditions.
Based on these results we proposed a combination that yields most accurate results and is suitable to be applied on human specimens. 
Last, we demonstrated that a dataset captured with the proposed method is suitable as a benchmark to compare different methods for 3D surface reconstruction.
\section{Acknowledgments}
This work has been supported by the OR-X - a swiss national research infrastructure for translational surgery - and associated funding by the University of Zurich and University Hospital Balgrist, and the Swiss Center for Musculoskeletal Imaging (SCMI). We thank Manuel Saladin for his contributions to the CT segmentation work.

\newpage

\newpage
\section*{Appendix} 
\appendix
\renewcommand{\thesection}{\Alph{section}}
\section{Ellipse Detection}
\subsection{Spherical Marker Estimation}\label{ssec:spherical_marker_estimation}

\begin{figure}[htbp]
    \centering
    \begin{minipage}{0.48\textwidth}
        \centering
        \includegraphics[width=\textwidth]{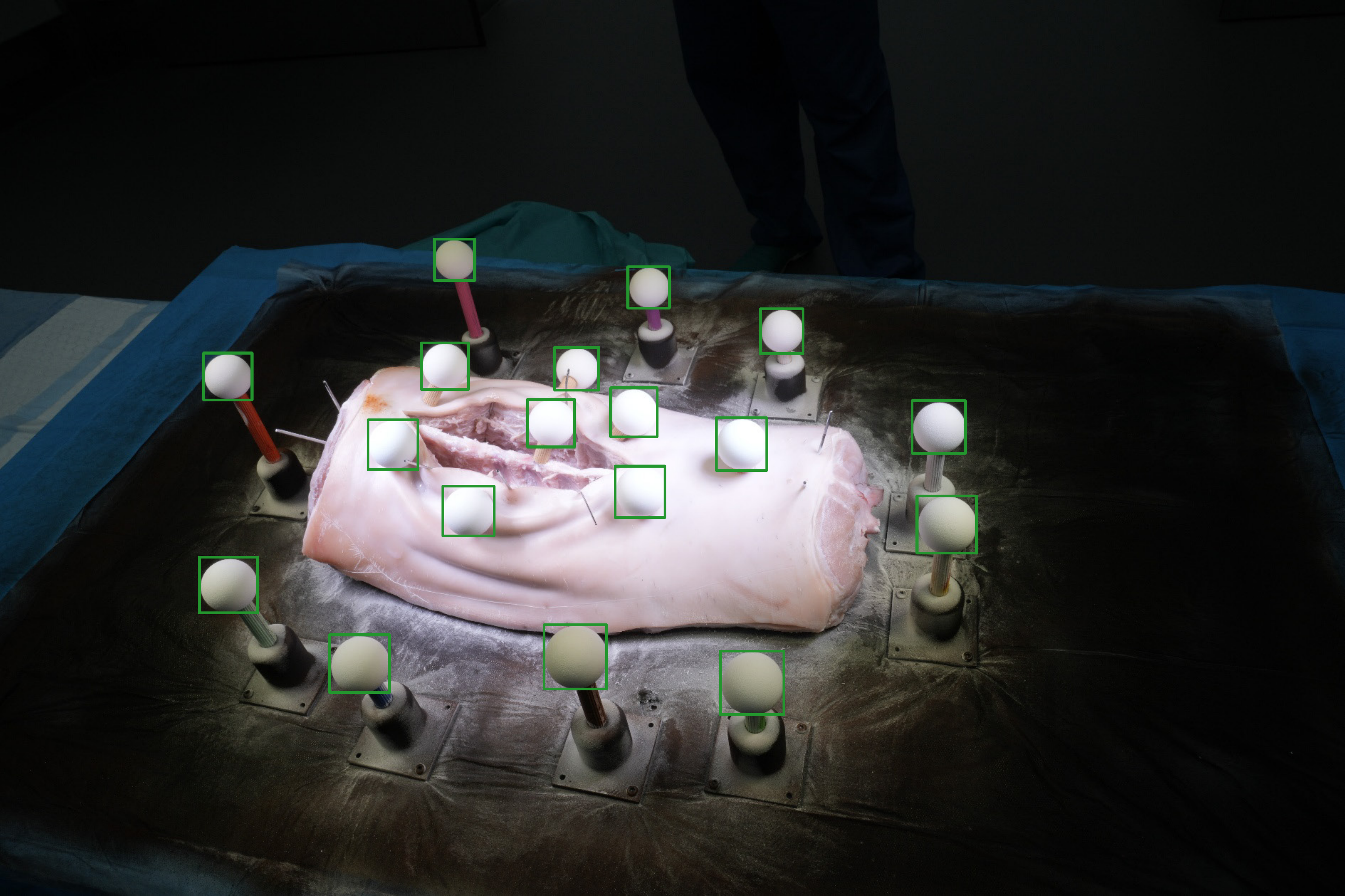}
    \end{minipage}
    \begin{minipage}{0.48\textwidth}
        \centering
        \includegraphics[width=\textwidth]
        {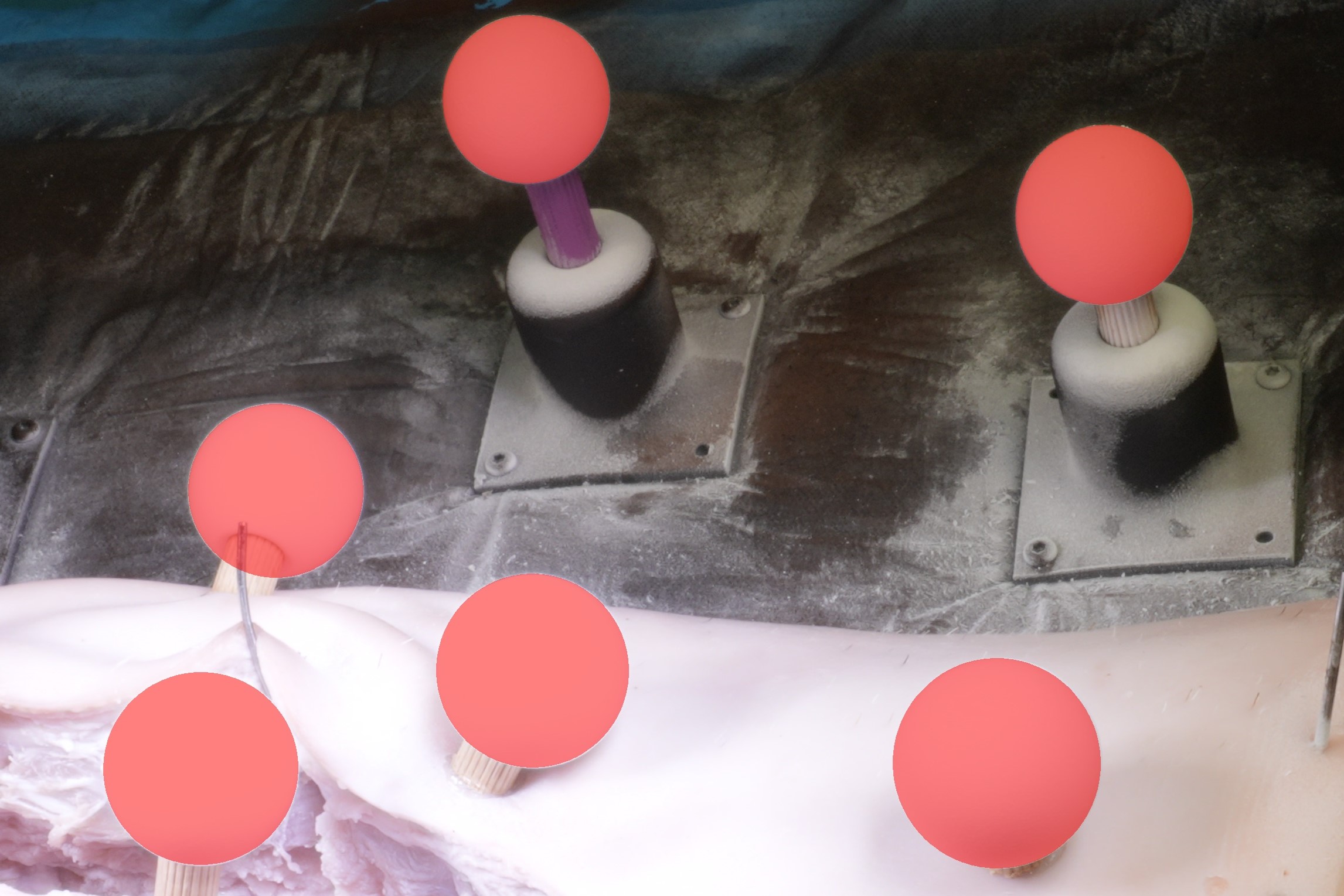}
    \end{minipage}
    \caption{Bounding boxes are extracted for all markers from a high-resolution image (left). This serves as a basis for the segmentation and fitting of ellipses on the segmented masks (right).}
    \label{fig:marker_estimation}
\end{figure}

To get a first estimation of the spherical markers used for scene registration, we perform a bounding box detection, followed by segmentation and ellipse estimation from the obtained masks. The process is illustrated in \cref{fig:marker_estimation}.

\textit{Sphere Detection} Sphere detection is performed using Grounding DINO \cite{liu2025grounding}, a model designed to detect arbitrary objects based on text inputs. We utilize the prompt ``spheres'' to find bounding boxes around the spheres in the image. If fewer spheres are detected than expected, we lower the confidence threshold to capture more uncertain detections, ensuring that the number of detected spheres meets a minimum count.

\textit{Sphere Segmentation} Once bounding boxes are obtained, SAM2 \cite{ravi2024sam2} is used to segment the spheres within these regions. Starting from the center of each bounding box (assumed to be within the sphere), SAM2 produces a binary mask \( m_c \in \{0, 1\}^{w \times h} \) for each color channel \(c\in\{R,G,B\}\), where $w\times h$ are the respective widths and heights of the bounding boxes. 

These masks are aggregated across channels by computing $m_\text{agg}=m_\text{R}\lor m_\text{G}\lor m_\text{B}$ and then postprocessing the resulting mask using morphological operations such as erosion and dilation. Only the connected component containing the center pixel \(\left(\lfloor w/2\rfloor, \lfloor h/2\rfloor\right)\) is retained. 

\textit{Ellipse Fitting} Finally, ellipse fitting is performed to approximate the spheres, which appear as ellipses when projected to two dimensions. The ellipse parameters \(\left(x_\text{center}, y_\text{center}, a, b, \theta\right)\), where \((x_\text{center}, y_\text{center})\) is the center, \(a\) and \(b\) are the semi-major and semi-minor axes, and \(\theta\) is the rotation angle, are estimated by minimizing a loss function. This loss function \(\mathcal{L}\) is defined as the sum of false positives (FP) and false negatives (FN), where a FP is defined as \(\mathbf{e}_{ij} = 1 \land \mathbf{m}_{ij} = 0\), and a FN as \(\mathbf{e}_{ij} = 0 \land \mathbf{m}_{ij} = 1\), with \(\mathbf{e}\) being the binary mask of the fitted ellipse.

To ensure better convergence and avoid local optima, the regression is repeated with multiple starting conditions. The ellipse parameters that result in the lowest loss are chosen as the final estimate. If more ellipses are detected than expected, the ellipses with the highest losses are discarded, leaving the desired number of spheres. The final set of ellipse parameters provides a rough estimate, serving as a starting point for further pixel-level detection, described in Sec. 3.2 of the main manuscript.

\subsection{Extraction of Accurate Ellipse Edge Points}
\begin{figure}[h]
    \centering
    \includegraphics[width=\linewidth]{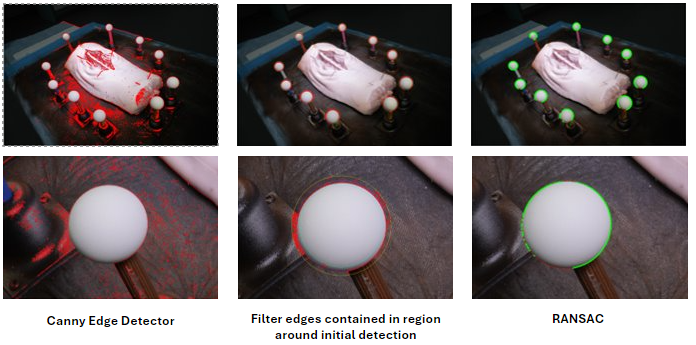}
    \caption{Extraction of accurate ellipse edge points, from left to right}
    \label{fig:enter-label}
\end{figure}
After obtaining an initial marker estimation, we detect Canny edges in the images and keep only the edge points contained in an envelope around the initial estimation. Then, we perform RANSAC to filter outliers and samples points equidistantly along the ellipse that corresponds to the output of RANSAC. In our experiments, we sampled K=200 points.

\rev{\section{Camera and End-Effector Calibration}}
The camera intrinsics were recovered using the MATLAB Computer Vision Toolbox, assuming a standard pinhole camera model with two radial and two tangential distortion parameters. For calibration, we captured 90 high-resolution images of a professional checkerboard pattern\footnote{\url{https://calib.io/}} and obtain a mean reprojection error of \rev{0.42} px for a resolution of 9504 x 6336. Throughout the calibration process and subsequent experiments, the focal length was fixed, and the aperture was set to its minimum (f/22) to maximize the depth of field. Shutter speed and ISO values were manually adjusted at the start to ensure proper exposure during data acquisition (ISO 100 and 1/15 sec in our case).

\rev{The robot's end-effector to camera calibration was performed following an approach similar to \cite{ozguner2020}.}

\rev{\section{Marker design}
In this section, we detail the design of the markers used for both scene registration and evaluation of the ground truth accuracy.}

\begin{figure}
    \centering
    \includegraphics[width=0.7\linewidth]{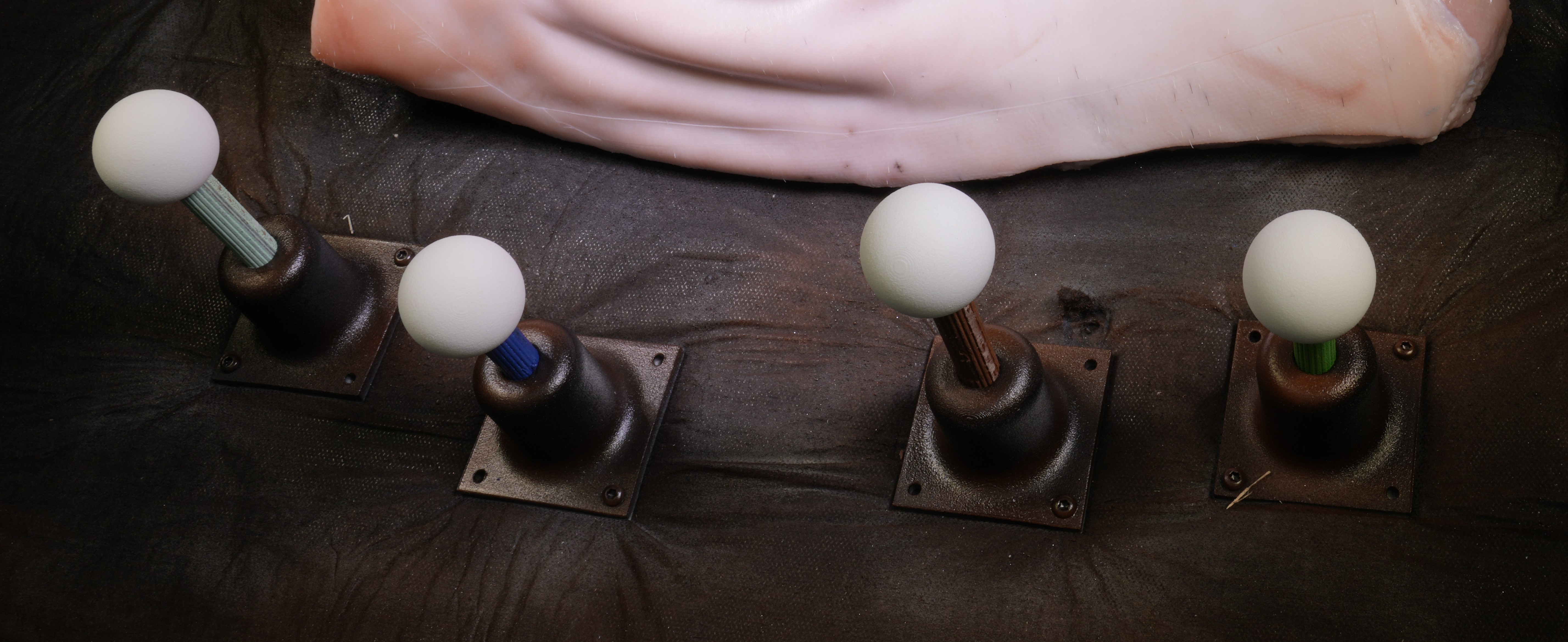}
    \caption{Up-close view of four markers used for scene registration.}
    \label{fig:marker_examples}
\end{figure}

\rev{We employ 3D markers consisting of spheres mounted on rigid wooden cylinders, each a few centimeters high and vertically fixed onto a wooden board where the specimen is placed. Both the spheres and the attachment bases of the wooden cylinders are 3D-printed.
This design is chosen for its flexibility and precision, allowing easy adjustment of the markers' location and height to suit the size and shape of the specimen. These adjustments improve marker visibility within the camera’s field of view and help reduce occlusions caused by the specimen. Figure \ref{fig:marker_examples} provides an close-up view of the markers affixed to the wooden board adjacent to the anatomy during data capture.}

\rev{\section{Evaluation}}

\noindent \rev{All quantitative evaluations are carried out using Chamfer distance between the reconstructed mesh $\mathcal{P}$ and the ground truth one $\mathcal{G}$.
For a reconstructed point $\hat{\mathbf{x}} \in \mathcal{P}$, its distance to the ground truth is defined
as follows:}
\rev{\begin{equation}
\label{eq:dist}
    d_{\hat{\mathbf{x}}\rightarrow \mathcal{G}} = \underset{\mathbf{x}\in \mathcal{G}}{\operatorname{min}}
\| \hat{\mathbf{x}} - \mathbf{x} \|,
\end{equation}}
\rev{and vice versa for a ground truth point $\mathbf{x} \in \mathcal{G}$ and its distance to the reconstructed mesh.}
\rev{The distance measures are accumulated over the entire meshes to define the Chamfer distance} 
\rev{\begin{equation}
    CD =
    \frac{1}{2}\left ( \frac{1}{|\mathcal{P}|} \sum_{\hat{\mathbf{x}} \in \mathcal{P}} d_{\hat{\mathbf{x}}\rightarrow \mathcal{G}}
    +
    \frac{1}{|\mathcal{G}|} \sum_{\mathbf{x} \in \mathcal{G}} d_{\mathbf{x}\rightarrow \mathcal{P}} \right )
\end{equation}}

\rev{To minimize sampling error in the distance measurement (Eq. \ref{eq:dist}), the meshes were upsampled, maintaining a point spacing of 0.1 mm to adequately represent the surface geometry. Distances exceeding 20 mm were classified as outliers and excluded from the final score calculation similarly to \cite{dtu_sp}.
}

\begin{figure}[h]
    \centering
    \includegraphics[width=0.7\linewidth]{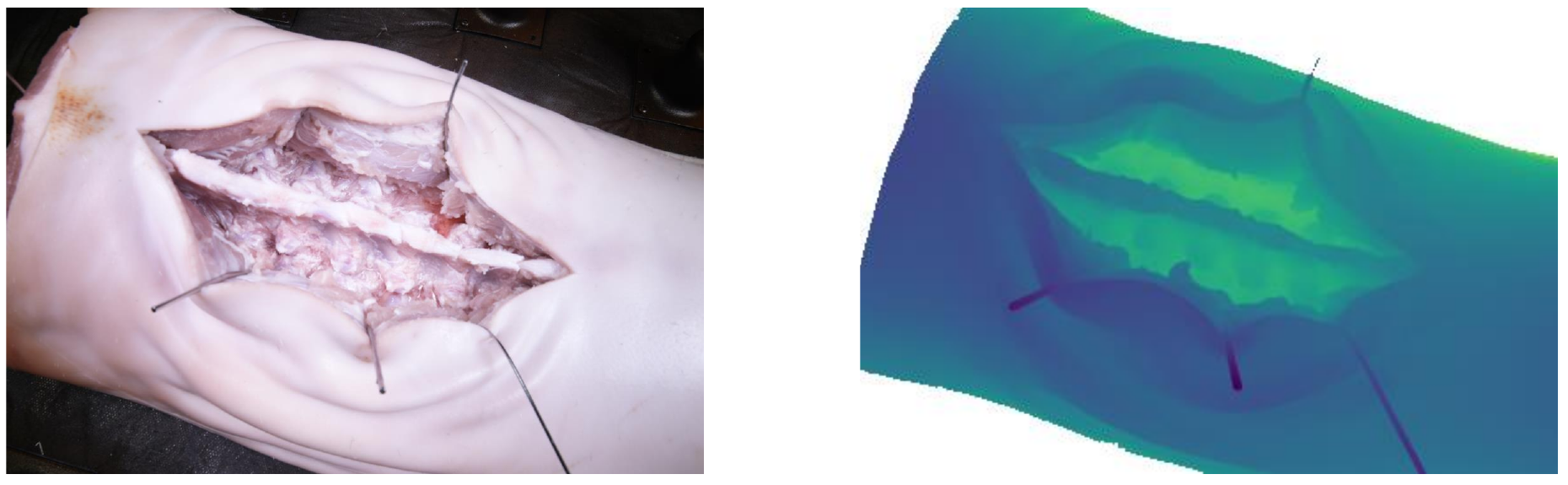}
    \caption{\rev{\textbf{Left:} A sample of a typical image in our dataset, obtained with our method. It has a resolution of 9504x6636 pixels. \textbf{Right:} The associated depth map derived from our ground truth mesh.}}
    \label{fig:sample_without_markers}
\end{figure}

\begin{figure}
    \centering
    \includegraphics[width=1\linewidth]{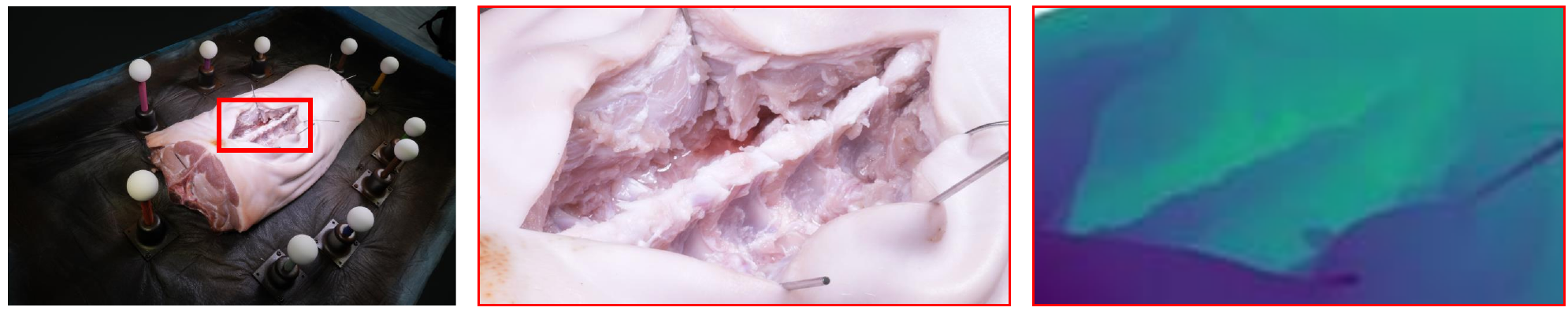}
    \caption{\rev{
    \textbf{Left:} Raw image of resolution 9504x6336 px. 
    \textbf{Middle:} Cropped image of resolution 1920x1080 px which corresponds to a Full HD resolution.
    \textbf{Right:} Depth image derived from our ground truth mesh.}}
    \label{fig:marker_cropping}
\end{figure}

\rev{\section{Dataset Samples}}
\rev{Figure \ref{fig:sample_without_markers} presents a sample image from our dataset with its associated depth map.
For images containing markers, these can be removed by cropping  while preserving high resolution. The resulting marker-free images remain accurately associated with the 3D ground truth, as demonstrated in Figure \ref{fig:marker_cropping}.}

\end{document}